\documentclass[10pt]{article}
\usepackage[utf8]{inputenc}
\usepackage[T1]{fontenc}
\usepackage{arxiv}
\usepackage{authblk}
\usepackage{graphicx}
\usepackage{amsmath}
\usepackage{amssymb}
\usepackage{bm}
\usepackage{url}
\usepackage[capitalise]{cleveref}
\usepackage{xcolor}
\usepackage{colortbl}
\usepackage{array}
\usepackage{listings}

\title{GPU-accelerated simulated annealing based on p-bits with real-world device-variability modeling}

\author[1]{*Naoya Onizawa}
\author[1]{Takahiro Hanyu}
\affil[1]{Research Institute of Electrical Communication, Tohoku University, Sendai, 980-8577, Japan}

\begin{document}
	
	\flushbottom
	\maketitle

\begin{abstract}
	Probabilistic computing using probabilistic bits (p-bits) presents an efficient alternative to traditional CMOS logic for complex problem-solving, including simulated annealing and machine learning. Realizing p-bits with emerging devices such as magnetic tunnel junctions (MTJs) introduces device variability, which was expected to negatively impact computational performance. However, this study reveals an unexpected finding: device variability can not only degrade but also enhance algorithm performance, particularly by leveraging timing variability. This paper introduces a GPU-accelerated, open-source simulated annealing framework based on p-bits that models key device variability factors—timing, intensity, and offset—to reflect real-world device behavior. Through CUDA-based simulations, our approach achieves a two-order magnitude speedup over CPU implementations on the MAX-CUT benchmark with problem sizes ranging from 800 to 20,000 nodes. By providing a scalable and accessible tool, this framework aims to advance research in probabilistic computing, enabling optimization applications in diverse fields.
\end{abstract}
	%
	%
	\thispagestyle{empty}

\section*{Introduction}

In recent years, a new computational model called the probabilistic bit (p-bit) has emerged \cite{IL}.
Unlike traditional bits, which are either 0 or 1, a p-bit can take any state between 0 and 1, with a probability distribution. 
p-bits can be implemented through software \cite{p-bit_emulation} or hardware such as Magnetoresistive Random Access Memory (MRAM)  \cite{p-bit_device} or FPGAs \cite{IL_FPGA,CIL,p-bit_FPGA,p-bit_async_impl}.
They are particularly effective in Boltzmann machines \cite{Boltzmann1984}, invertible logic \cite{CIL_training}, and applications like Bayesian inference \cite{p-bit_BI}, parallel tempering \cite{p-bit_PT}, Gibbs sampling \cite{p-bit_gibbs}, and simulated annealing (SA) \cite{p-bit_general, SSA, SSQA}.
SA is a stochastic optimization method \cite{SA1,SA2} used for solving combinatorial problems like MAX-CUT \cite{SA_max-cut} and enhancing machine learning algorithms \cite{SA_ML}.
These problems, often represented by Ising models, are computationally challenging (NP-hard)  \cite{NP-hard}.
SA aims to minimize the `energy' of the system, which represents the cost function of the problem. 
Hardware implementations of SA have also been explored for large-scale problems \cite{DA,Ising_PT,JETCAS_SSA}.
Other methods for solving Ising models include coherent Ising machines \cite{CIM}, simulated bifurcation \cite{SB}, and coupled oscillation networks \cite{SA_oscillation}. 
Quantum annealing (QA) promises faster solutions \cite{QA1,QA2} but is currently limited by device performance \cite{QA_GI,QA_review}.

SA using p-bits (pSA) is based on a probabilistic computing approach, allowing it to be implemented on conventional computers. 
This concept positions pSA as a promising method for tackling large-scale problems more efficiently. 
One of the key advantages of pSA is its ability to update multiple nodes simultaneously, unlike traditional SA, which updates nodes sequentially. 
This parallel update capability could accelerate the process of reaching the global minimum energy state, thereby speeding up the search for the optimal solution. 
Early research has shown that pSA performs well on small-scale problems \cite{p-bit_general}. 
However, as the problem size increases, pSA's effectiveness tends to decline. 
Simulations have revealed that for larger problems, such as graph isomorphism \cite{GI} and MAX-CUT, pSA struggles to maintain its efficiency \cite{SSA}.
In particular, the energy of the Ising model fails to decrease as expected, suggesting that pSA faces difficulties in finding optimal or near-optimal solutions for larger-scale problems.

Recently, new algorithms for pSA have been proposed to address these challenges, showing that optimal solutions can be attained even for large-scale problems \cite{TApSA}. 
These innovations have made significant strides in overcoming the limitations seen in earlier research, enhancing the overall performance of pSA in solving complex optimization tasks more efficiently. 
The introduction of these algorithms has opened up the possibility of large-scale implementations of pSA, which marks a major step forward in the practical application of probabilistic computing. 
However, achieving this goal requires careful consideration of the inherent variability in p-bit devices. 
Unlike traditional deterministic systems, p-bits are realized through probabilistic devices such as magnetic tunnel junction (MTJ) devices \cite{p-bit_device_fast1,p-bit_device_fast2}, which are subject to variations in their behavior. 
Recent studies have highlighted that implementations of MTJ devices, commonly used in MRAM, exhibit significant variability across different parameters \cite{p-bit_variability}. 
This variability presents a critical challenge in the practical deployment of pSA at scale, as device-level inconsistencies can impact the overall performance and reliability of the algorithm. 
As such, the validation and optimization of pSA algorithms in large-scale implementations will require simulations that rigorously account for these variations. 
Developing accurate models that incorporate the effects of device variability is essential for understanding how pSA performs in real-world scenarios and for ensuring that large-scale p-bit systems can reliably solve complex problems.

 In this article, we present a novel pSA simulator that incorporates device variability, addressing a critical challenge for future large-scale p-bit implementations. 
 Specifically, the simulator models key types of variability, including `timing' (device speed), `intensity' (response to input signals), and `offset' (shifts in input signals). 
 By accounting for these factors, the simulator provides a more accurate representation of real-world performance in pSA systems, which is essential for assessing the viability of large-scale implementations.
 To handle the computational demands of simulating under various conditions of device variability, we have implemented the simulator using CUDA \cite{CUDA} for GPU acceleration. 
This allows us to efficiently simulate complex systems with numerous variability scenarios, achieving a significant two-order-of-magnitude speedup over traditional CPU-based simulations.
Using the NVIDIA RTX 4090 GPU, we observed this performance boost specifically in benchmarks involving the MAX-CUT problem, demonstrating the effectiveness of GPU acceleration for such complex optimization tasks.
  The ability to run high-speed simulations makes it feasible to explore a wide range of conditions and configurations, which would be impractical using slower methods.
 Actually, we observed that device variability, especially timing variability, can lead to improved performance.
 In addition to its performance benefits, one of the key contributions of this work is the development of an open-source tool that can be used by the broader research community.

  \section*{Methods}

	\subsection*{p-bits based simulated annealing (pSA)}
	
	\paragraph{p-bits and pSA.}
	The intrinsic probabilistic nature of p-bits allows them to serve as a powerful tool for solving certain problem types that leverage randomness or uncertainty. 
	The output state of a p-bit can be described as follows:
	\begin{equation}
		\sigma_i(t+1) = {\rm sgn}\Bigl(r_i(t) + {\rm tanh}\bigl(I_i(t)\bigr)\Bigr), 
		\label{eqn:pbits}
	\end{equation}
	where $\sigma_i(t+1) \in \{-1,1\}$ is a binary output signal, $I_i(t)$ is a real-valued input signal, and $r_i(t) \in \{-1:1\}$ is a random signal.
	%
	The signal $r_i(t)$ is based on the p-bit model, where it is assumed to follow a uniform random distribution \cite{IL}. 
	This assumption aligns with the theoretical formulation of p-bits, where randomness is uniformly distributed to ensure unbiased stochastic behavior.
	Furthermore, when implementing p-bits using hardware devices such as MTJs, the design ensures that the generated random signal closely follows a uniform distribution. 
	This design principle guarantees consistency between theoretical models and hardware implementations.
	\color{black}
	
	\begin{figure}[t]
		\centering
		\includegraphics[width=0.8\linewidth]{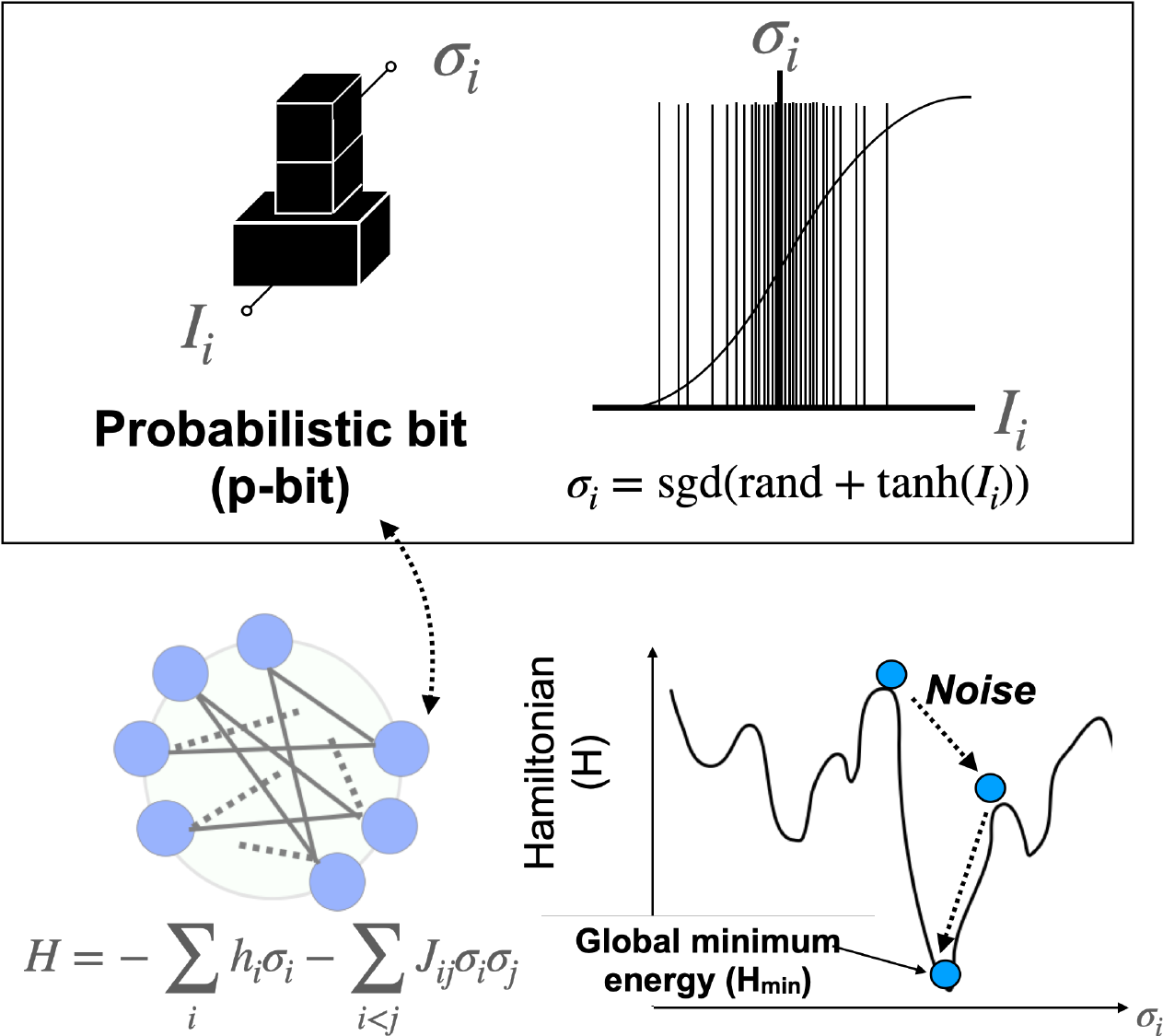}
		\caption{
			Simulated annealing using p-bits (pSA) operates based on the probabilistic nature of p-bits (top), as described by \cref{eqn:pbits}. A combinatorial optimization problem is mapped onto an Ising model, which corresponds to an energy function (Hamiltonian). In this model, each p-bit is biased by $h$ and interacts with other p-bits through weights $J$ (bottom left).  pSA seeks to reduce the energy of the Ising model by altering the states of p-bits $\sigma_i$. When the global minimum energy $H_{min}$ is reached, the states $\sigma_i$ represent a solution to the combinatorial optimization problem (bottom right).
		}
		\label{fig:pSA}
	\end{figure}
	
	The p-bit-based simulated annealing (pSA) method \cite{p-bit_general} is illustrated in \cref{fig:pSA}.
	A combinatorial optimization problem is modeled using the Ising model, which represents the system's energy. 
	This energy is governed by a Hamiltonian, expressed as follows:
	\begin{equation}
		H(\sigma) = - \sum_i h_i\sigma_i - \sum_{i < j} J_{ij}\sigma_i\sigma_j,
		\label{eqn:Ising}
	\end{equation}
	where $\sigma_i \in \{-1,1\}$ represents a binary state, $h$ denotes the biases applied to p-bits, and $J$ signifies the connection weights between p-bits. 
	Depending on the specific combinatorial optimization problem, different values of $h$ and $J$ are assigned.
	 In pSA, each p-bit is affected by a bias $h$ and interacts with other p-bits through the weights $J$.
    The input to each p-bit, $I_i(t)$, is calculated based on the outputs of other p-bits and is given by the following expression:
	\begin{equation}
		I_i(t) = I_0 \left (h_i+\sum_j J_{ij}\cdot \sigma_j(t) \right),
		\label{eqn:conv}
	\end{equation}
	where $I_0$ is a pseudo inverse temperature used to control the simulated annealing.
	
	Simulated annealing seeks to find the global minimum energy of \cref{eqn:Ising} by adjusting the states of $\sigma_i$. 
	The specific method for updating $\sigma_i$ varies depending on the SA algorithm employed \cite{SA1,p-bit_general,Ising_PT,SSA,SSQA}. 
	In pSA, the pseudo inverse temperature $I_0$ is gradually increased to lower the energy of the Ising model. 
	When $I_0$ is low, $\sigma_i$ can freely flip between `-1' and `+1', enabling the algorithm to explore various potential solutions to the combinatorial optimization problem. 
	As $I_0$ rises, $\sigma_i$ stabilizes, steering the system toward the global minimum energy. 
	At this stage, the values of $\sigma_i$ represent the solution to the combinatorial optimization problem.

	\paragraph{Variations of pSA.}
	Two new pSA algorithms have been developed to tackle the challenges associated with applying pSA to large-scale combinatorial optimization problems \cite{TApSA}. 
	The first of these is the time-averaged p-bit simulated annealing algorithm (TApSA), which incorporates a time-average operation into the pSA framework. 
	This operation, as introduced in \cref{eqn:conv} of the pSA algorithm, is defined as follows:
	\begin{subequations}
		\begin{equation}
			TI_i(t) =h_i+\sum_j J_{ij}\cdot \sigma_j(t) ,
			\label{eqn:TApSA1}
		\end{equation}
		\begin{equation}
			I_i(t) =  I_0 \left (\frac{1}{\alpha}\sum_{i=0}^{\alpha-1}TI_i(t-i)\right),
			\label{eqn:TApSA2}
		\end{equation}
		\label{eqn:TApSA}
	\end{subequations}
	where, $TI_i(t)$ represents a temporary value used in the time-averaging operation, and $\alpha$ denotes the size of the time window over which $TI_i(t)$ is averaged. 
	The input to the p-bit, $I_i(t)$, as defined in \cref{eqn:pbits}, is also applied in TApSA. 
	The equations in \cref{eqn:TApSA} compute the time-averaged p-bit input signal. 
	This averaging operation smooths the signal over a specified time window, effectively reducing random fluctuations or `noise' in the signal.

	The second algorithm is simulated annealing based on stalled p-bits (SpSA). 
	In SpSA, the input of a p-bit, denoted as $I_i(t)$, is probabilistically stalled, retaining the value of $I_i(t-1)$ from the previous time step. 
	The equation for SpSA, as described earlier, is given by:
	\begin{equation}
		I_i(t) = \begin{cases} 
			I_i(t-1), & \text{with probability of getting stalled } p \\
			I_0 \Bigg(h_i+\sum_j J_{ij}\cdot \sigma_j(t)\Bigg). & \text{with probability} (1-p) 
		\end{cases}
		\label{eqn:SpSA}
	\end{equation}
	In this equation, the input of the p-bit at time $t$, $I_i(t)$, can either be stalled, meaning it remains the same as the input from the previous time step $I_i(t-1)$ with probability $p$, or it can take on a new value with probability $(1-p)$. 
	This approach represents a significant departure from conventional pSA, where \cref{eqn:conv} is replaced by \cref{eqn:SpSA} in the SpSA algorithm.

	\subsection*{Modeling p-bits with device variability}

	\begin{figure}[t]
		\centering
		\includegraphics[width=1.0\linewidth]{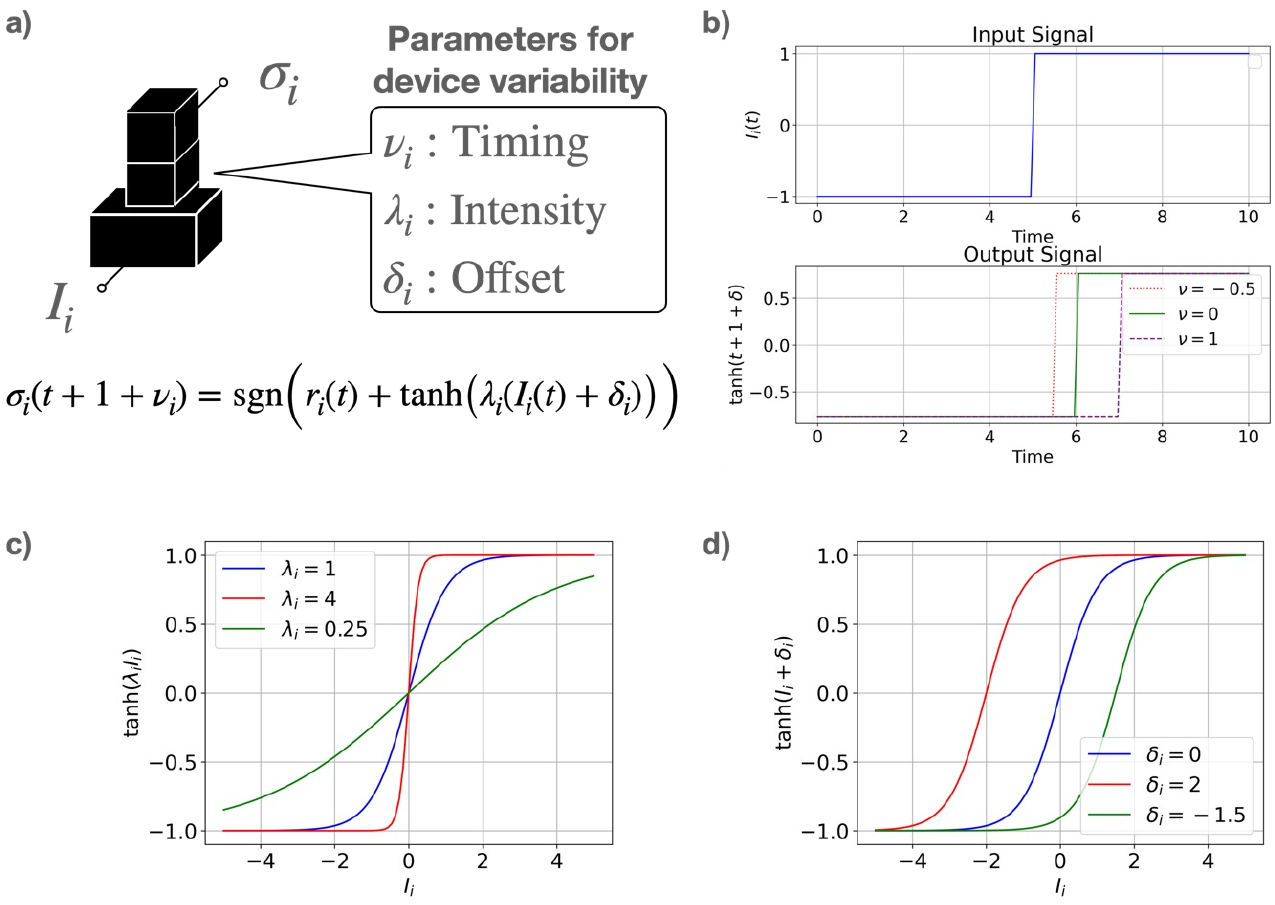}
		\caption{Modeling of p-bits with device variability. The model introduces three new parameters to capture device variability: timing ($\nu_i$), intensity ($\lambda_i$), and offset ($\delta_i$). (a) These parameters accurately reflect the switching characteristics of each MTJ, enabling the model to account for variations in p-bit outputs due to changes in the MTJ dwell time. (b) The timing parameter $\nu_i$ captures shifts in the output signal in response to input signal changes. (c) The intensity parameter $\lambda_i$ captures the variability in the steepness of the $\tanh(\lambda_i I_i)$ function, indicating intensity to input changes. (d) The offset parameter $\delta_i$ captures shifts in the input threshold, which affect the transition points of the output curve. These parameters combined allow the model to reflect probabilistic behavior variations in p-bits according to the properties of each MTJ.}
		\label{fig:p-bit_v}
	\end{figure}

p-bits are realized through probabilistic devices such as MTJs \cite{p-bit_device_fast1,p-bit_device_fast2}. 
MTJs consist of two ferromagnetic layers separated by an insulating barrier. 
The relative orientation of the magnetizations in these layers determines the resistance of the MTJ. 
When the magnetizations are parallel, the resistance is low (parallel state), and when they are antiparallel, the resistance is high (antiparallel state). 
This change in resistance is measured as Tunnel Magnetoresistance (TMR), which is a key parameter in the performance of p-bits.

Recently, the observed variations in MTJs primarily relate to the stochastic switching behavior between these parallel and antiparallel states, which in turn affects the performance of p-bit devices \cite{p-bit_variability}. 
Specifically, the following variations were highlighted:
\begin{itemize} 
\item \textbf{TMR:} 
TMR represents the ratio between the resistances in the parallel and antiparallel states of an MTJ. It was observed that too high a TMR value leads to undesirable plateaus in the p-bit output, where the device becomes ``stuck" in one state for too long. On the other hand, too low a TMR value reduces the range of fluctuation, limiting the ability of the p-bit to explore various probabilistic states. The ideal range for TMR was found to balance between these extremes, providing a sufficient fluctuation range without causing plateaus. 
\item \textbf{Dwell Times:} 
Dwell time refers to the duration the MTJ spends in either the parallel or antiparallel resistance state before switching. This is another important source of variation. Faster MTJs, characterized by shorter dwell times, tend to produce more continuous resistance distributions, leading to smoother p-bit outputs. In contrast, slower MTJs with longer dwell times are more prone to exhibit a bimodal distribution, which can result in plateaus in the p-bit's probabilistic output.
\end{itemize}
	To include the device variability of p-bits,  the output state of the $i$-th p-bit is updated from \cref{eqn:pbits} as follows:
		\begin{equation}
		\sigma_i(t+1+\nu_i) = {\rm sgn}\Bigl(r_i(t) + {\rm tanh}\bigl(\lambda_i(I_i(t)+\delta_i)\bigr)\Bigr), 
		\label{eqn:pbits_variability}
	\end{equation}
	where three new parameters of $\lambda_i$, $\delta_i$, and $\nu_i$ are included as shown in \cref{fig:p-bit_v}
	\begin{itemize}
	 \item \textbf{Intensity} 
	 $\lambda_i$ is a scaling parameter that adjusts the impact of the input current $I_i(t)$ on the p-bit's output.
	A smaller $\lambda_i$ reduces the  of the p-bit output to the input current, making the output fluctuations more prominent. 
	This parameter is connected to the TMR  variations of the MTJ.
	 Large TMR variability can cause $\lambda_i$ to vary, altering the switching intensity of the p-bit.
	\item \textbf{Offset} 
	$\delta_i$ is a bias shift or offset parameter that controls how much the input current $I_i(t)$ deviates from its true value.
	$\delta_i$ can be influenced by MTJ resistance state variations (parallel and antiparallel states) or device-level offsets. For example, variability due to manufacturing processes or temperature changes can cause $\delta_i$ to vary, thereby shifting the switching probability threshold.
	\item \textbf{Timing}
	$\nu_i$ represents the delay parameter that controls when the next state update occurs.
	$\nu_i$ is associated with the average dwell time of the MTJ or the switching speed. Faster MTJs (with shorter dwell times) have a smaller $\nu_i$, resulting in quicker state updates, while slower MTJs (with longer dwell times) will have a larger $\nu_i$, indicating a delayed state update.
	Containing $\nu_i$ results in the need for a simulation at finer intervals than a single cycle in \cref{eqn:pbits_variability}.
	This implies that the state $\sigma_i$ must be updated at time steps smaller than the duration of one cycle. 
	We refer to this finer subdivision of time as the time resolution of the simulation, $T_{res}$. 
	By increasing the time resolution, we are able to more accurately capture the effects of device variations, thereby improving the precision of the simulation results.
	
	\end{itemize}
	$\lambda_i $and $\delta_i$ should capture the TMR and energy barrier variability accurately so that the model can reflect how each p-bit’s probabilistic characteristics change according to the MTJ properties.
    $\nu_i$ should adequately represent the switching speed of each MTJ so that the model can capture the timing variations in p-bit outputs due to MTJ dwell time variations.
    %
    
   The three parameters representing the device variability are modeled by normal distributions as follows:
    \begin{subequations}
    	\begin{align}
    		\lambda_i &\sim \mathcal{N}(1, \sigma_{\lambda}^2) \label{eq:lambda} \\
    		\delta_i &\sim \mathcal{N}(0, \sigma_{\delta}^2) \label{eq:delta} \\
    		\nu_i &\sim \mathcal{N}(0, \sigma_{\nu}^2) \label{eq:nu}
    	\end{align}
    	\label{eqn:variability}
    \end{subequations}
    Varying the standard deviation allows simulations under various conditions of variability.
    \color{black}
 
	\subsection*{CUDA programming and experimental conditions}

	The p-bit with the device variability is modeled in CUDA (Compute Unified Device Architecture) \cite{CUDA} for large-scale p-bit computing.
    CUDA is a software framework and application programming interface (API) that allows developers to use graphics processing units (GPUs) for general-purpose processing.
	The pseudo code of  the p-bit with the device variability is described as follows:
	\begin{lstlisting}[language=Python, caption={Pseudo code of the SimulatedAnnealingModule that updates p-bits based on device variability in CUDA. The function iteratively updates the state of p-bits using a probabilistic rule involving hyperparameters such as \textit{lambda}, \textit{delta}, and \textit{nu}.}]
	
	Function SimulatedAnnealingModule(vertex, mem_I0, h_vector, J_matrix, spin_vector, rnd, lambda, delta, nu, count_device):
		
		For each index i in range(0, vertex):
			If (count_device mod nu[i] == 0):
				Initialize D_res with h_vector[i]
		
				For each index k in range(0, vertex):
					D_res = D_res + J_matrix[i][k] * spin_vector[k]
		
				Itanh = tanh(lambda[i] * mem_I0 * (D_res + delta[i])) + rnd[i]
		
				If (Itanh > 0):
					spin_vector[i] = 1
				Else:
					spin_vector[i] = -1
	\end{lstlisting}
     This pseudo code presents a simulated annealing module that models the behavior of p-bits using variability parameters in a CUDA environment. 
     The function \texttt{SimulatedAnnealingModule} takes in several inputs, including the number of nodes (\texttt{vertex}), the pseudo inverse temperature, $I_0$ (\texttt{mem\_I0}), and coupling matrices (\texttt{J\_matrix}). 
          It also considers the variability parameters such as \texttt{lambda}, \texttt{delta}, and \texttt{nu}, along with a random number vector (\texttt{rnd}).
          %
     The random number vector \texttt{rnd} is updated at every simulation cycle. This ensures that each probabilistic update of the spin vector is based on a new set of random values, maintaining the stochastic nature of the simulated annealing process.  
     \color{black}
     This CUDA code is controlled from Python using PyCUDA 
     \cite{PyCUDA}  

     \begin{table}[h]
     	\centering
     	\begin{tabular}{|c|c|c|c|c|}
     		\hline
     		\rowcolor{gray!50}
     		Graph  & \# nodes & Structure & Weights ($J$) & \# edges\\
     		\hline
     		G1 &800 & random  & {+1} &19176\\
     		\hline
     		G22 &  2000 & random & {+1} &19990 \\
     		\hline
     		G47&  1000 & random  &{+1} &9990  \\
     		\hline
     		G48 &  3000 & troidal & {+1, -1}&6000 \\
     		\hline
     		G55 & 5000 & random & {+1}  &12498  \\
     		\hline
     		G60& 7000	 & random & {+1} & 17148\\
     		\hline
     		G67	& 10000	& troidal & {+1, -1} & 20000\\
     		\hline
     		G77	& 14000	& troidal & {+1, -1} & 28000\\
     		\hline
     		G81	& 20000	& troidal & {+1, -1} & 40000\\
     		\hline
     	\end{tabular}
     	\caption{
     		Summary of MAX-CUT benchmarks used for evaluating simulated annealing based on p-bits with device variability. The graphs labeled as Gxx are part of the G-set benchmark set. Each graph varies in the number of nodes, structure type, edge weights ($J$), and total number of edges. These benchmarks were selected to explore the effect of variability on large-scale problem instances.
     	}
     	\label{tb:graph}
     \end{table}
     
     The simulated annealing based on the p-bit with  the variability is evaluated  in MAX-CUT problems.
     The p-bit based annealing  algorithms of pSA, TApSA, and SpSA are simulated.
     pSA, TApSA, and SpSA compute \cref{eqn:conv}, \cref{eqn:TApSA}, and \cref{eqn:SpSA}, respectively, with the device-variability p-bit model of \cref{eqn:pbits_variability}.
     The simulations are carried out using Python 3.6 on Intel Xeon Gold 6430 and NVIDIA RTX 4090 with 128 GB of memory.
     %
     In our current GPU implementation, we use 32 threads per block, where the total number of blocks is defined as $\lceil \text{number of nodes} / 32 \rceil$.
      This configuration was chosen based on empirical testing to balance computational efficiency and memory access patterns.
     The relationship between thread count and GPU performance is not always linear, and increasing the number of threads does not necessarily guarantee better performance. 
     While 32 threads have provided a good balance in our testing, we recognize that further investigation into dynamic thread allocation based on problem size could potentially improve GPU utilization. This remains an area of future exploration.
     \color{black}

     The MAX-CUT problem aims to partition a graph into two groups such that the sum of the weights of the edges connecting the two groups is maximized. 
     This process involves cutting' the graph into two separate sections, hence the term ``MAX-CUT''.
     The MAX-CUT benchmark, namely G-set, is used for simulations (\cref{tb:graph}), where the G-set includes Gxx graphs that vary in node sizes and edge connections \cite{G-set}.
     A crucial component in these pSA algorithms is the manipulation of the pseudo-inverse temperature, which plays a significant role in guiding the annealing process through the solution space. 
     During the simulated annealing process, the pseudo-inverse temperature $I_0$ is gradually increased over time, starting from an initial value $I_{0min}$ and reaching a maximum value $I_{0max}$, following the formula $I_0(t+1) = I_0(t)/\beta$.
     %
     At smaller values of $I_0(t)$, the p-bit states are more prone to flipping, allowing for broad exploration of the solution space. As the value grows larger, the p-bit states become more stable, facilitating convergence to an optimal solution. This annealing process is central to ensuring the system transitions smoothly from exploration to exploitation.
     \color{black}
     The hyperparameters for the simulated annealing processes, such as $I_{0min}$, $I_{0max}$, and $\beta$, are not arbitrarily selected. 
     Rather, they are determined using a specific statistical method designed to optimize the performance of the simulated annealing algorithm (SSA) \cite{Hyperparameter_SSA}.
     Similar to the previous study \cite{TApSA}, in this simulation, $I_{0min}$ is set to $\frac{0.1}{{\rm mean}(s_i)}$ and $I_{0max}$ is set to $\frac{10}{{\rm mean}(s_i)}$, where $s_i$ is defined as $\sqrt{(n-1)\cdot {\rm Var}(J_{i,:})}$. 
     The parameter $\beta$ is calculated as $\Bigl(\frac{I_{0min}}{I_{0max}}\Bigr)^{\frac{1}{cycle-1}}$.
     Here, $n$ is the number of nodes in the graph, $cycle$ is the number of cycles from $I_{0min}$ to $I_{0max}$, and $J_{i,:}$ is a vector containing all the edge weights connected to the $i$-th p-bit.
    For TApSA and SpSA,  based on the results \cite{TApSA}, the values $\alpha = 4$ and $p = 0.5$ are selected for \cref{eqn:TApSA} and \cref{eqn:SpSA}, respectively.
     %
     In the studies where these algorithms were originally proposed, the parameter values were determined through extensive sweeps across a range of possible values to identify optimal settings for various problem instances. While the optimal parameter values can vary depending on the specific problem being solved, both alpha and p generally exhibit robust performance when set above certain threshold values, making them effective across a wide range of problems.  
     $\alpha = 4$ and $p = 0.5$ were chosen as they have been shown to perform well across diverse problem sets and provide a reasonable balance between exploration and convergence.  
     \color{black}
     When simulating the device variability, $T_{res}$ is set to 10 for the timing variability, $\nu$.

      To evaluate the simulation speed of the GPU implementation, a CPU version was developed using Python and executed on the same machine as the GPU implementation.
      In our current CPU implementation, the CPU version does not utilize parallel processing across multiple CPU cores. The performance baseline was evaluated using a single-threaded implementation in Python. Although Python provides libraries such as multiprocessing and concurrent.futures for parallel execution, enabling efficient multi-core utilization would require restructuring the CPU code, especially for tasks involving shared memory or frequent data exchange.
     \color{black}

\section*{Results}

 \subsection*{Speed comparison of CPU and GPU}

 \begin{figure}[t!]
 	\centering
 	\includegraphics[width=1.0\linewidth]{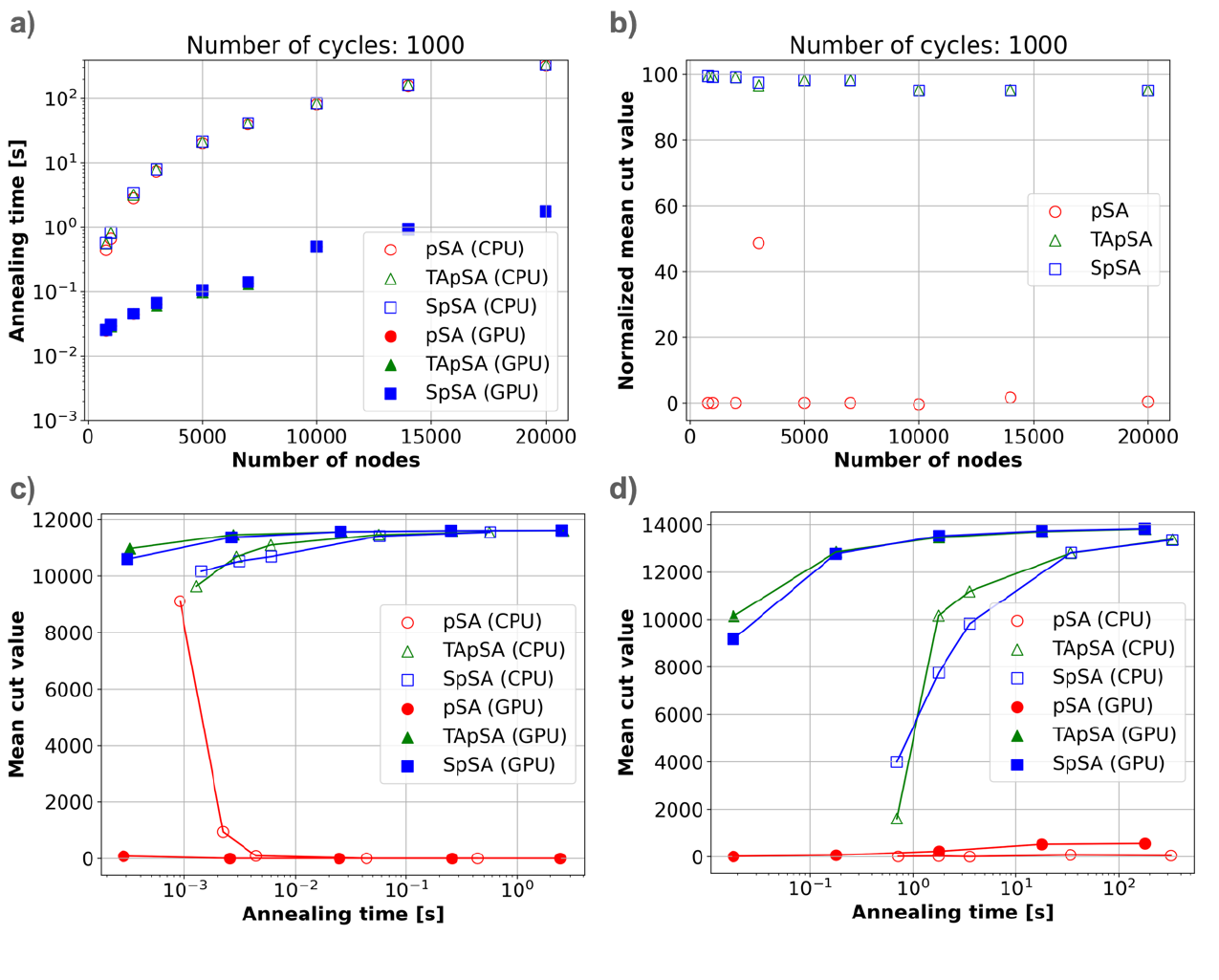}
 	\caption{Performance comparison of pSA, TApSA, and SpSA algorithms on CPU and GPU across different problem sizes without the device variability.
 		(a) Annealing time as a function of the number of nodes: GPU implementations show significantly faster annealing times compared to their CPU counterparts, while pSA (GPU) exhibits longer times for larger problem sizes.
 		(b) Normalized mean cut value as a function of the number of nodes: The SpSA and TApSA algorithms maintain high normalized mean cut values across all problem sizes, indicating their robustness in achieving optimal solutions.
 		(c) Mean cut value versus annealing time for different algorithms in G1: As the annealing time increases, TApSA and SpSA converge to higher mean cut values, particularly in the GPU implementation. pSA struggles to find higher cut values efficiently on both CPU and GPU.
 		(d) Extended analysis of mean cut value versus annealing time in G81: The GPU-based implementations of TApSA and SpSA consistently achieve higher mean cut values faster than their CPU implementations. In contrast, pSA on the GPU lags in performance relative to the other algorithms.}
 	\label{fig:without}
 	\vspace{-3mm}
 \end{figure}
 
 \cref{fig:without} demonstrates the comparative performance of three simulated annealing-based algorithms—pSA, TApSA, and SpSA —executed on both CPU and GPU for a range of problem sizes without the device variability.
 Subplot (a) shows the scaling behavior of annealing time against the number of nodes for each algorithm.
 All the benchmarks in \cref{tb:graph} are used with 1000 cycles. 
 It is evident that GPU-based versions  achieve significantly lower annealing times compared to their CPU counterparts, indicating the efficiency gains obtained from parallel computation. 
 Subplot (b) illustrates the normalized mean cut values across different node sizes. 
The normalized mean cut value is obtained by calculating the ratio of the average cut value to the best-known cut value, with the average cut value derived from 100 trials.
 Both SpSA and TApSA maintain a high normalized mean cut value across all problem sizes on both CPU and GPU, highlighting their ability to find near-optimal solutions consistently. 
On the other hand, pSA shows a significant drop in cut value, indicating reduced effectiveness compared to the other two algorithms, as reported in the previous study \cite{TApSA}.
Subplots (c) and (d) further analyze the mean cut value as a function of annealing time for G1 and G81, respectively, offering insights into the convergence behavior of each algorithm.
 The annealing time is increased when the number of cycles is increased.
 Subplot (c) shows that SpSA and TApSA quickly converge to higher mean cut values on GPU, reflecting their faster solution convergence with increased annealing time. 
 Subplot (d) extends this analysis, reinforcing the observation that TApSA and SpSA efficiently leverage the GPU to achieve higher cut values, while pSA's GPU implementation struggles to compete in both efficiency and solution quality.
 These results underscore the effectiveness of TApSA and SpSA, particularly on GPU, in solving large-scale problems efficiently, achieving near-optimal solutions with shorter annealing times.
 
 \subsection*{Simulation analysis of device variability}

\begin{figure}[h!]
	\centering
	\includegraphics[width=1.0\linewidth]{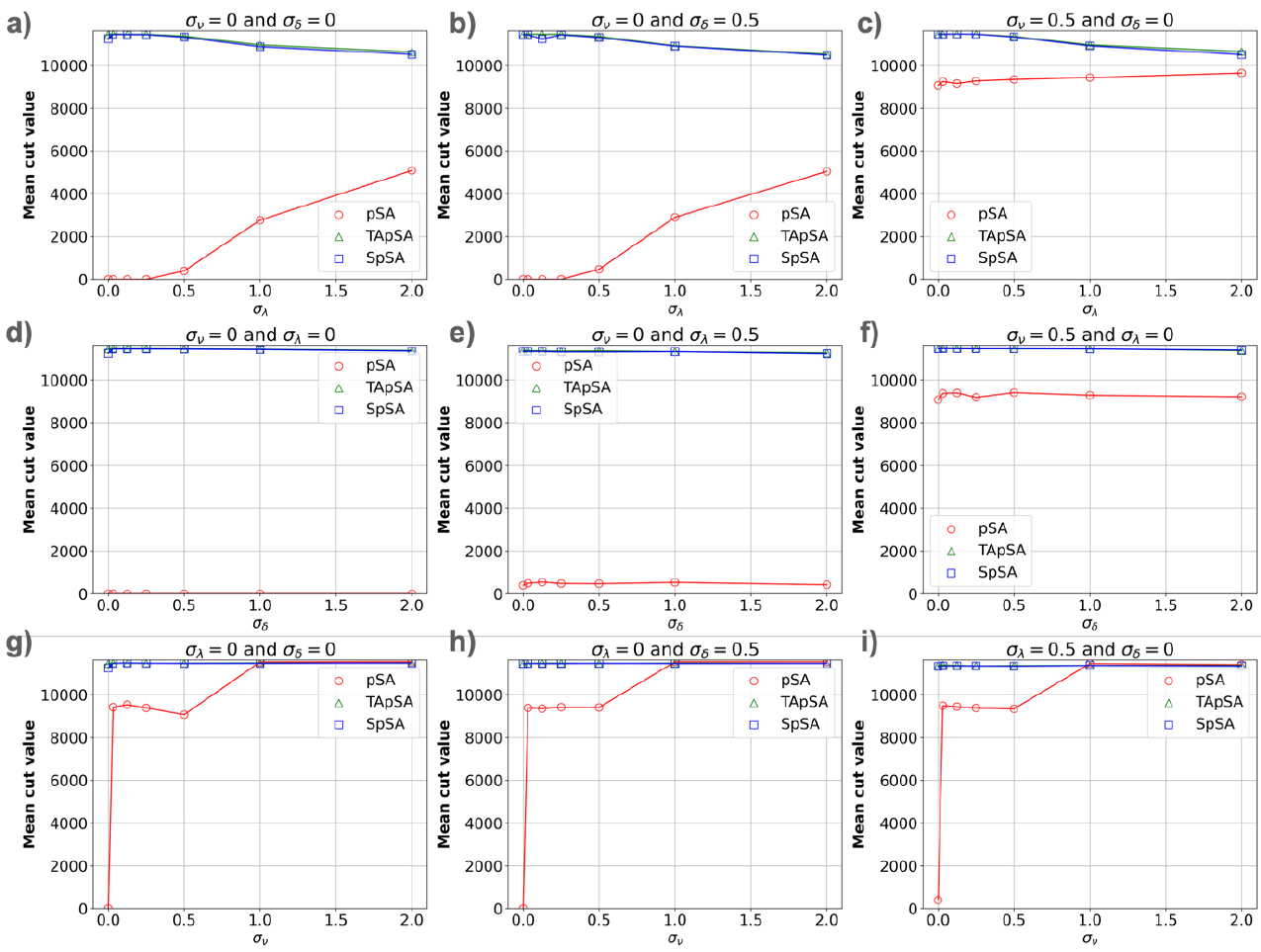}
	\caption{Impact of variability parameters on mean cut value for different simulated annealing algorithms for G1. 
		Each subplot shows the sensitivity analysis of three algorithms (pSA, TApSA, and SpSA) under variations in key variability parameters: $\sigma_{\lambda}$, $\sigma_{\delta}$, and $\sigma_{\nu}$.
		(a), (b), and (c): Effects of $\sigma_{\lambda}$ under different combinations of $\sigma_{\delta}$ and $\sigma_{\nu}$. When both $\sigma_{\delta} = 0$ and $\sigma_{\nu} = 0$, the mean cut values for pSA improve steadily with increasing $\sigma_{\lambda}$, while TApSA and SpSA maintain consistently high values (a). As $\sigma_{\delta}$ increases to 0.5 (b), the effect on pSA’s mean cut values diminishes slightly. The trends remain stable when $\sigma_{\nu}$ increases to 0.5 (c), showing resilience in TApSA and SpSA. Notably, in case (c), pSA also achieves a high mean cut value, indicating its robustness under these conditions.
		(d), (e), and (f): Effects of $\sigma_{\delta}$ under different combinations of $\sigma_{\lambda}$ and $\sigma_{\nu}$. With $\sigma_{\lambda} = 0$ and $\sigma_{\nu} = 0$ (d), the pSA shows a gradual improvement in mean cut values with increasing $\sigma_{\delta}$, while TApSA and SpSA maintain stable high performance. When $\sigma_{\lambda}$ increases to 0.5 (e), the patterns remain similar. As $\sigma_{\nu}$ increases to 0.5 (f), the stability in SpSA and TApSA continues.
		(g), (h), and (i): Effects of $\sigma_{\nu}$ under different combinations of $\sigma_{\lambda}$ and $\sigma_{\delta}$. For $\sigma_{\lambda} = 0$ and $\sigma_{\delta} = 0$ (g), the impact on pSA is evident, with an initial increase in mean cut values. As $\sigma_{\lambda}$ increases to 0.5 (h), the stability of TApSA and SpSA remains unaffected, while pSA shows a slight decrease in mean cut values. When $\sigma_{\delta}$ increases to 0.5 (i), all algorithms demonstrate high stability, with SpSA consistently performing the best.}
	\label{fig:cut_value_G1}
\end{figure}

\cref{fig:cut_value_G1} analyzes the effects of variability parameters $\sigma_{\lambda}$, $\sigma_{\delta}$, and $\sigma_{\nu}$ for G1 on the performance of three simulated annealing algorithms: pSA, TApSA, and SpSA.
The variability parameters defined by \cref{eqn:variability} can change the variability of p-bits for the input-signal intensity ($\lambda$), the input-signal offset ($\delta$), and the update timing of p-bits ($\nu$). 
The mean cut value is plotted against increasing values of each variability parameter, while the other two parameters are fixed at specific values.
Each row of subplots presents a different variability parameter being varied, highlighting how the algorithms respond to changes in $\sigma_{\lambda}$, $\sigma_{\delta}$, and $\sigma_{\nu}$. The key observations are:
\begin{itemize}
	\item \textbf{Robustness of TApSA and SpSA:} Across all parameter variations, TApSA and SpSA demonstrate resilience to changes in variability, maintaining consistently high mean cut values. This suggests that these algorithms are effective in achieving near-optimal solutions even under variable device conditions.
	\item \textbf{Sensitivity of pSA:} The pSA algorithm shows a clear sensitivity to changes in all three variability parameters, with a noticeable improvement in mean cut values as the parameters increase. It was confirmed that pSA performance benefits from device variability, with a particularly significant impact observed from \(\sigma_{\nu}\), which represents the variability in the timing of p-bit updates. This variability implies that updates are not carried out synchronously but rather in a somewhat asynchronous manner. This de-synchronization appears to positively influence pSA performance, although it still generally falls short of TApSA and SpSA in most cases.
\end{itemize}
These findings underscore the advantage of using TApSA and SpSA in applications where device variability plays a critical role, as these algorithms exhibit stability and effectiveness under varying conditions.

\begin{figure}[t]
	\centering
	\includegraphics[width=1.0\linewidth]{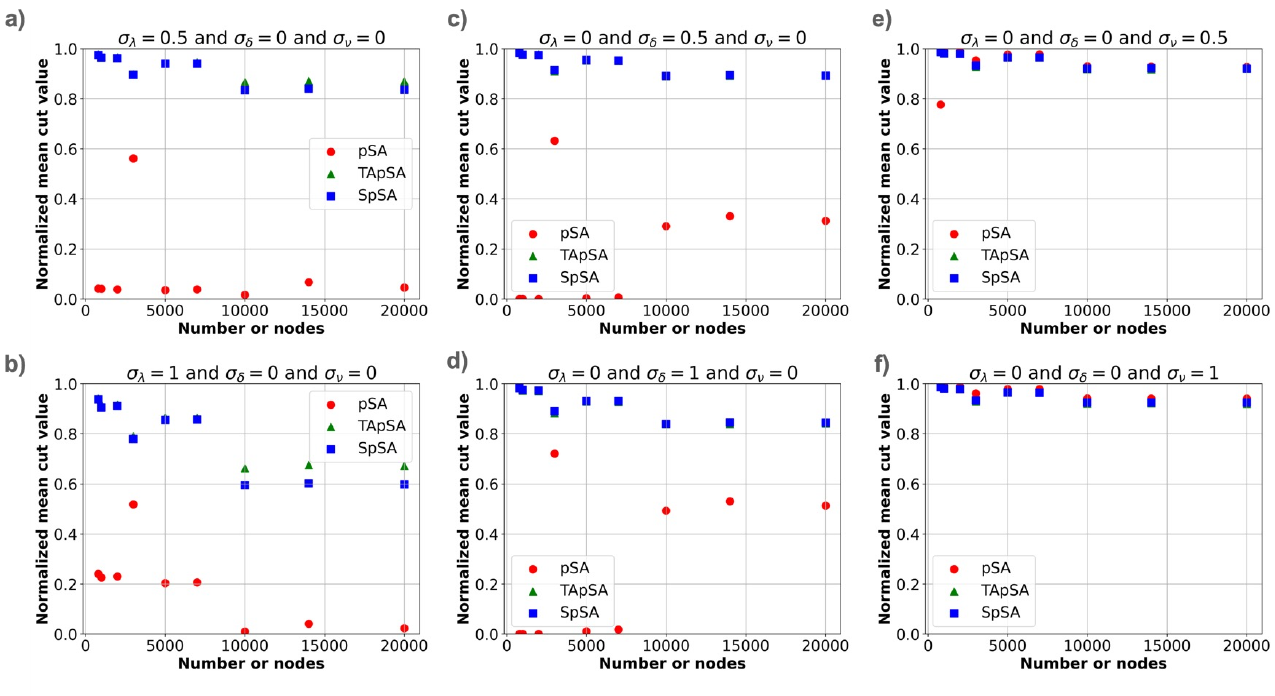}
	\caption{Impact of variability in $\sigma_{\lambda}$, $\sigma_{\delta}$, and $\sigma_{\nu}$ on the normalized mean cut value for different node sizes. Each subplot shows how the normalized mean cut value changes with increasing number of nodes under various levels of variability for pSA, TApSA, and SpSA.
		(a): $\sigma_{\lambda} = 0.5$, $\sigma_{\delta} = 0$, and $\sigma_{\nu} = 0$: SpSA and TApSA maintain high normalized mean cut values across all node sizes, while pSA shows lower cut values with a significant drop at higher node counts.
		(b): $\sigma_{\lambda} = 1$, $\sigma_{\delta} = 0$, and $\sigma_{\nu} = 0$: As $\sigma_{\lambda}$ increases to 1, SpSA continues to perform well, but TApSA shows a small decrease in performance, and pSA remains low.
		(c): $\sigma_{\lambda} = 0$, $\sigma_{\delta} = 0.5$, and $\sigma_{\nu} = 0$: Variability in $\sigma_{\delta}$ affects pSA more strongly, reducing its cut values, while TApSA and SpSA remain largely unaffected.
		(d): $\sigma_{\lambda} = 0$, $\sigma_{\delta} = 1$, and $\sigma_{\nu} = 0$: Further increase in $\sigma_{\delta}$ results in little to no effect on SpSA and TApSA, while pSA still performs poorly.
		(e): $\sigma_{\lambda} = 0$, $\sigma_{\delta} = 0$, and $\sigma_{\nu} = 0.5$: Introducing variability in $\sigma_{\nu}$ shows resilience in both SpSA and TApSA across all node sizes, but pSA remains significantly affected.
		(f): $\sigma_{\lambda} = 0$, $\sigma_{\delta} = 0$, and $\sigma_{\nu} = 1$: Even under high variability in $\sigma_{\nu}$, SpSA and TApSA maintain high normalized cut values, whereas pSA remains sensitive and continues to increase the normalized mean cut values.}
	\label{fig:node}
\end{figure}

\cref{fig:node} illustrates the effects of varying $\sigma_{\lambda}$, $\sigma_{\delta}$, and $\sigma_{\nu}$ on the normalized mean cut values for three different algorithms—pSA, TApSA, and SpSA—across different node sizes. The normalized mean cut value reflects the quality of the solution relative to the optimal, with a value of 1 representing an optimal cut.
\begin{itemize}
	\item \textbf{Resilience of SpSA and TApSA:} Across all six subplots, SpSA and TApSA demonstrate robustness to variability in $\sigma_{\lambda}$, $\sigma_{\delta}$, and $\sigma_{\nu}$, consistently achieving high normalized mean cut values regardless of node size. This indicates that these algorithms are well-suited for solving large-scale problems under a range of device conditions.
	\item \textbf{Sensitivity of pSA:} The pSA algorithm shows a much higher sensitivity to changes in the variability parameters. In all cases, pSA achieves lower normalized mean cut values, especially as the number of nodes increases. This suggests that pSA is less capable of finding optimal or near-optimal solutions when subject to device variability, particularly when the problem size becomes large.
\end{itemize}
These results highlight the superior performance of SpSA and TApSA under varying device conditions, making them more reliable choices for large-scale optimization tasks where device variability may impact the solution quality.

\section*{Discussion}

\begin{figure}[t]
	\centering
	\includegraphics[width=1.0\linewidth]{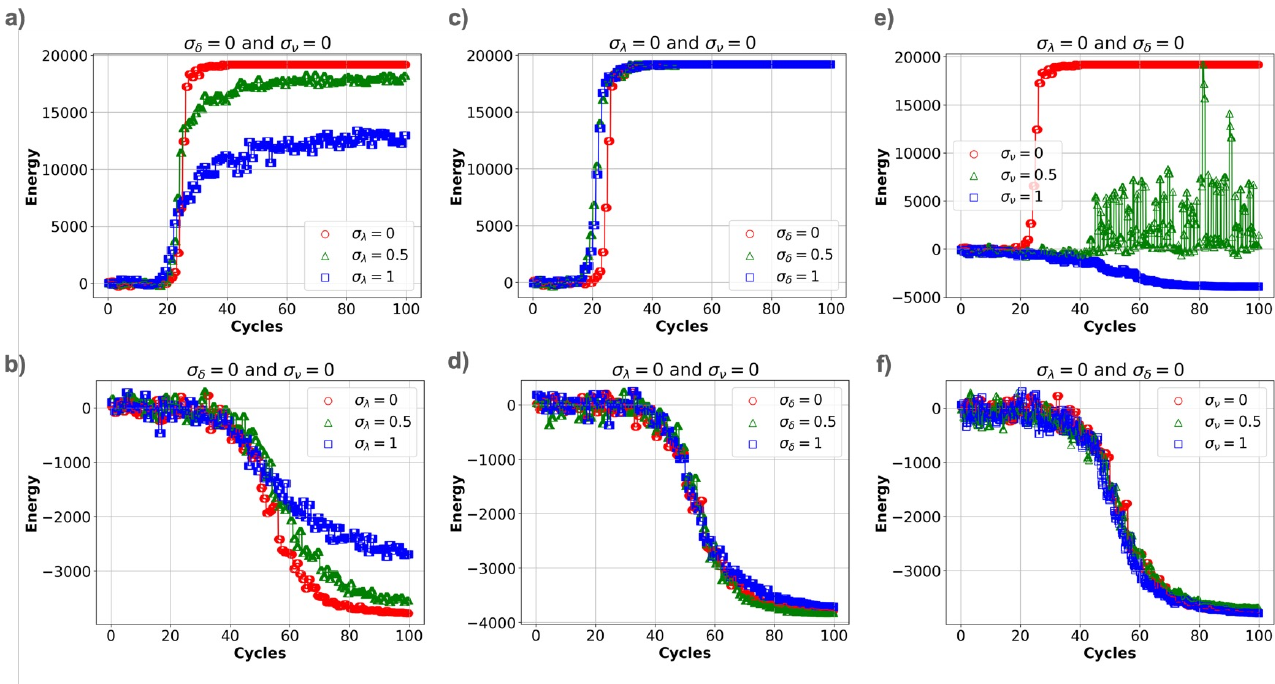}
	\caption{Energy evolution across simulated annealing cycles under varying levels of $\sigma_{\lambda}$, $\sigma_{\delta}$, and $\sigma_{\nu}$ for G1. The subplots  (a), (c), and (e) are obtained by pSA and the others are obtained by TApSA. Each subplot shows the energy trajectory over 100 cycles with different variability conditions applied to specific parameters.
		(a): $\sigma_{\delta} = 0$, $\sigma_{\nu} = 0$, and varying $\sigma_{\lambda}$ values (0, 0.5, 1). Higher $\sigma_{\lambda}$ values lead to elevated final energy levels, indicating improved performance in finding low-energy solutions as variability in $\sigma_{\lambda}$ increases.
		(b): $\sigma_{\delta} = 0$, $\sigma_{\nu} = 0$, and varying $\sigma_{\lambda}$, focusing on systems optimized to minimize energy. Increased $\sigma_{\lambda}$ results in slower convergence toward lower energy states, with less effective minimization as $\sigma_{\lambda}$ rises.
		(c): $\sigma_{\lambda} = 0$, $\sigma_{\nu} = 0$, and varying $\sigma_{\delta}$ (0, 0.5, 1). In (c), the results show minimal impact from variability in $\sigma_{\delta}$, with little effect on the final energy values.
		(d): $\sigma_{\lambda} = 0$, $\sigma_{\nu} = 0$, focusing on energy minimization with varying $\sigma_{\delta}$. The results show that as $\sigma_{\delta}$ increases, the system's ability to reach low-energy states slightly diminishes.
		(e): $\sigma_{\lambda} = 0$, $\sigma_{\delta} = 0$, with varying $\sigma_{\nu}$ (0, 0.5, 1). At $\sigma_{\nu} = 0.5$, energy behavior is unstable, while at $\sigma_{\nu} = 1$, energy decreases steadily, reaching a near-optimal state.
		(f): $\sigma_{\lambda} = 0$, $\sigma_{\delta} = 0$, focusing on energy minimization with varying $\sigma_{\nu}$. The results show minimal impact from changes in $\sigma_{\nu}$, with no significant difference in convergence performance.}
	\label{fig:energy}
\end{figure}

To investigate in detail how device variability influences normalized mean cut values, we examined the energy changes during annealing.
\cref{fig:energy}  presents the impact of varying device variability parameters ($\sigma_{\lambda}$, $\sigma_{\delta}$, and $\sigma_{\nu}$) on energy evolution during simulated annealing cycles for G1. 
The subplots (a), (c), and (e) are obtained by pSA, and the others are obtained by TApSA. 
Since the performance of TApSA and SpSA is nearly identical, only TApSA was observed in this analysis.
Each subplot explores how different variability conditions affect the annealing process, which aims to reach a stable, low-energy configuration.
\begin{itemize}
	\item \textbf{Effect of $\sigma_{\lambda}$:} In plots (a) and (b), as $\sigma_{\lambda}$ increases, the system exhibits higher final energy states, indicating difficulty in finding low-energy solutions. In the minimization-focused plot (b), increased $\sigma_{\lambda}$ causes slower convergence and less effective energy minimization, emphasizing that higher $\sigma_{\lambda}$ disrupts stability.
	\item \textbf{Effect of $\sigma_{\delta}$:} Plots (c) and (d) show similar trends with $\sigma_{\delta}$. Higher variability in $\sigma_{\delta}$ has minimal impact on the final energy state and does not significantly affect annealing efficiency, as shown in (d) with varying $\sigma_{\delta}$ values.
	\item \textbf{Effect of $\sigma_{\nu}$:} In plots (e) and (f), increasing $\sigma_{\nu}$ in (e) makes it easier to reach low-energy states, indicating that greater variability in $\sigma_{\nu}$ can enhance convergence. In contrast, plot (f) shows that convergence to low-energy states is minimally affected by $\sigma_{\nu}$ variability, maintaining stable performance.
\end{itemize}
The analysis highlights how variability in parameters $\sigma_{\lambda}$, $\sigma_{\delta}$, and $\sigma_{\nu}$ impacts the annealing process. An increase in $\sigma_{\lambda}$ generally leads to higher final energy states and reduced convergence stability, making it challenging to reach low-energy solutions in TApSA. In contrast, $\sigma_{\delta}$ shows minimal effect on both energy states and annealing efficiency, indicating robustness against its variability in both pSA and TApSA. Interestingly, greater variability in $\sigma_{\nu}$ can facilitate convergence to low-energy states under certain conditions in pSA, suggesting that $\sigma_{\nu}$ variability may enhance the annealing process in some cases.

In our previous work, we observed that in the standard pSA implementation, p-bit states tended to oscillate, preventing the system from effectively reducing energy and reaching an optimal solution \cite{TApSA}. 
This issue likely arose because all p-bits were updated simultaneously, creating a synchronized behavior across the system that hindered proper state transitions.  
With the introduction of timing variability, the update timings of individual p-bits became slightly misaligned. This desynchronization appears to have mitigated the oscillation issue, allowing the system to explore the solution space more effectively and converge towards optimal or near-optimal solutions.  
It can be hypothesized that the slight asynchronicity introduced by timing variability plays a beneficial role in preventing undesired synchronization effects, thereby improving the overall efficiency of the SA process.  
\color{black}

In this study, we conducted a comprehensive performance analysis of three simulated annealing algorithms—pSA, TApSA, and SpSA—under varying device variability conditions. We evaluated the impact of variability parameters \(\sigma_{\lambda}\), \(\sigma_{\delta}\), and \(\sigma_{\nu}\) on annealing time, mean cut value, and energy evolution across different problem sizes.
Our findings show that SpSA and TApSA consistently outperform pSA, particularly for large-scale optimization problems, with resilience to changes in variability parameters. SpSA, in particular, demonstrates the highest efficiency and robustness, making it well-suited for real-world applications. In contrast, pSA is highly sensitive to variability, especially for larger problems, limiting its competitiveness.
Unexpectedly, we observed that device variability can both degrade and enhance algorithm performance. While it was anticipated that variability might lead to instability, particularly for pSA, we found that specific types of variability, such as timing variability, can actually improve algorithm efficiency in some cases. This dual effect highlights the complex role of device variability in p-bit-based simulated annealing and provides new insights into optimizing performance through controlled variability.
Additionally, we developed a novel pSA simulator that incorporates device variability to address challenges in large-scale p-bit implementations. Using CUDA for GPU acceleration, we achieved a significant speedup, enabling simulations under diverse variability conditions.
In energy evolution, SpSA and TApSA quickly converge to low-energy states, even under high variability, whereas pSA struggles with stability. This open-source simulator provides the research community with a valuable tool for exploring these algorithms and optimizing their performance for practical applications.

We would like to clarify that the main focus of this paper is not on benchmarking computational speed between the CPU and GPU implementations or comparing performance with alternative optimization algorithms. Instead, our primary objective is to demonstrate the impact of device variability (e.g., timing, parameter fluctuations) on p-bit implementations and to provide an open-source GPU-accelerated pSA simulator.
To validate our approach, we selected the MAX-CUT problem, a widely recognized benchmark for combinatorial optimization problems. While CPU-GPU speed comparisons are included as a supplementary analysis to illustrate the acceleration capabilities, they are not the central focus of our study.
We also recognize the importance of evaluating our simulator with other benchmark problems in future work to further validate its versatility and effectiveness across a broader range of combinatorial optimization challenges. Additionally, while a comparative analysis with other solvers could provide valuable insights, it falls outside the scope of this paper and may be considered as part of future research directions.
\color{black}

In terms of energy efficiency on optimization hardware, FPGA implementations of p-bit models would likely offer better energy efficiency compared to our GPU-based pSA simulator \cite{IL_FPGA}. However, our primary focus in this work is on developing a high-speed pSA simulator using GPUs to evaluate the impact of device variability. This simulator is intended as a precursor to hardware implementations that leverage future p-bit devices, such as MTJ-based devices.
If p-bit devices can be realized as actual hardware, it is expected that their energy efficiency would surpass that of traditional FPGA or ASIC implementations \cite{p-bit_device}. This is due to the inherently low-power nature of p-bit devices and their suitability for probabilistic computation. Consequently, while the current GPU implementation is not optimized for energy efficiency, it serves as a critical step toward evaluating and designing energy-efficient pSA hardware using real p-bit devices in the future.
\color{black}

	\section*{Data availability}
   All data generated or analyzed during this study are included in this published article. The Python codes are available at https://github.com/nonizawa/GPU-pSAv  \cite{nonizawa_GPU-pSAv}.
   
\bibliographystyle{naturemag}

	\section*{Acknowledgments}
	
	This work was supported in part by JST CREST Grant Number JPMJCR19K3, JSPS KAKENHI Grant Number JP21H03404, and  KIOXIA Corporation.
	
	\section*{Author contributions statement}
	N. O. conducted and analyzed the experiments. T. H. discussed the experiment. All authors reviewed the manuscript. 
	
	\section*{Additional information}
	Competing financial interests: The authors declare no competing financial interests.

\end{document}